\documentclass[10pt,twocolumn,letterpaper]{article}

\usepackage[pagenumbers]{cvpr} 

\usepackage{booktabs}
\usepackage{tabularx}

\definecolor{cvprblue}{rgb}{0.21,0.49,0.74}
\usepackage[pagebackref,breaklinks,colorlinks,allcolors=cvprblue]{hyperref}

\def\paperID{20972} 
\def\confName{CVPR}
\def\confYear{2026}

\title{MARVO: Marine-Adaptive Radiance-aware Visual Odometry}

\author{Sacchin Sundar\thanks{Equal contribution.}\\
University of Michigan\\
Ann Arbor, USA\\
{\tt\small sacchin@umich.edu}
\and
Atman Kikani\textsuperscript{*}\\
Vellore Institute of Technology\\
Chennai, India\\
{\tt\small atman.kikani2022@vitstudent.ac.in}
\and
Aaliya Alam\\
Vellore Institute of Technology\\
Chennai, India\\
{\tt\small aaliya.2022@vitstudent.ac.in}
\and
Sumukh Shrote\\
University of Pennsylvania\\
Philadelphia, USA\\
{\tt\small sshrote1@seas.upenn.edu}
\and
A. Nayeemulla Khan\\
Vellore Institute of Technology\\
Chennai, India\\
{\tt\small nayeemulla.khan@vit.ac.in}
\and
A. Shahina\\
Sri Sivasubramaniya Nadar College of Engineering\\
Tamil Nadu, India\\
{\tt\small shahinaa@ssn.edu.in}
}
\begin{document}
\maketitle
\begin{abstract}\
Underwater visual localization remains challenging due to wavelength-dependent attenuation, poor texture, and non-Gaussian sensor noise. We introduce MARVO, a physics-aware, learning-integrated Simultaneous Localization And Mapping framework that fuses underwater image formation modeling, differentiable matching, and reinforcement-learning optimization. At the front-end, we extend transformer-based feature matcher with a PhysicsAware Radiance Adapter that compensates for colorchannel attenuation and contrast loss, yielding geometrically consistent feature correspondences under turbidity. These semi dense matches are combined with inertial and pressure measurements inside a factor-graph localization backend, where we formulate a keyframe-based visual-inertial-barometric estimator using GTSAM library. Each keyframe introduces (i) Pre-integrated IMU motion factors, (ii) MARVO-derived visual pose factors, and (iii) barometric depth priors, giving a full-state MAP estimate in real time. Lastly, we introduce a Reinforcement-Learningbased Pose-Graph Optimizer that refines global trajectories beyond local minima of classical least-squares solvers by learning optimal retraction actions on SE(2).
\end{abstract}    
\section{Introduction}\
\label{sec:intro}
Advances in feature matching~\cite{loftr}, multi-sensor fusion~\cite{forster2017imu} and factor-graph optimization~\cite{dellaert2017factorgraphs} have empowered Visual Odometry (VO) and Simultaneous Localization and Mapping (SLAM) to achieve strong performance in terrestrial settings. Underwater environments, however, remain uniquely challenging. Light scattering, wavelength-dependent attenuation and strong non-Gaussian noise produce severe contrast loss, unstable features, and inconsistent long-horizon pose estimates. Classical VO and visual-inertial pipelines fail in such conditions because they neither correct the underlying physics of underwater image formation nor effectively fuse uncertain auxiliary sensors such as pressure and inertial measurements.

We propose the \textbf{MARVO}, a \textit{Marine-Adaptive Radiance-Aware Visual Odometry} framework that couples physics-guided front-end perception with probabilistic multi-sensor fusion and offline learning-driven pose-graph refinement. The basic idea of MARVO is that robust underwater VO calls for both (i) perception modules that explicitly compensate for radiometric distortions, and (ii) back-end optimization that can escape the local minima typical of noisy, visually degraded trajectories.

\begin{figure*}[t]
\centering
\includegraphics[width=\textwidth]{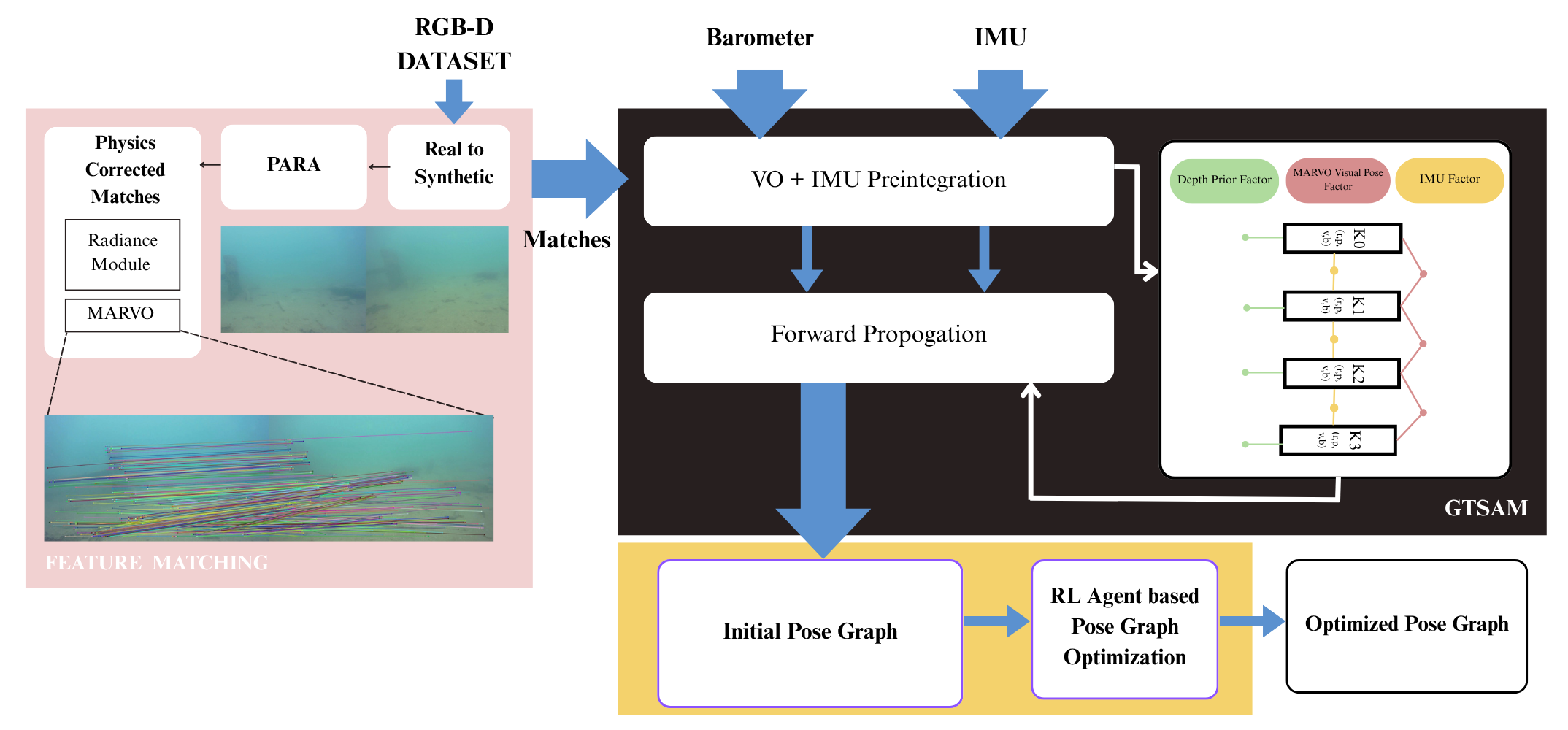}
\caption{Overview of MARVO. PARA enhances LoFTR features using physically-informed radiance correction. Corrected visual factors are fused with IMU and barometric depth in a GTSAM factor graph to produce real-time VO. An offline reinforcement-learning agent performs pose-graph refinement to obtain globally consistent trajectories.}
\label{fig:marvo_architecture}
\end{figure*}

At the front end, MARVO extends LoFTR~\cite{loftr} with a lightweight \textit{Physics-Aware Radiance Adapter (PARA)}. PARA modulates intermediate descriptors using learned attenuation and visibility estimates, counteracting color-channel imbalance and restoring feature matchability before transformer attention. This radiance-aware formulation enables stable semi-dense correspondences in regions where standard LoFTR degrades.

These corrected visual factors are fused together with IMU preintegration~\cite{forster2017imu} and barometric depth measures within a keyframe-based factor graph implemented in GTSAM~\cite{dellaert2017factorgraphs}. Each keyframe contributes with PARA-enhanced visual pose constraints, motion factors and unary depth priors to create a consistent maximum-a-posteriori estimate. The system acts like a fixed-lag smoother, preserving real-time performance while maintaining global consistency across heterogeneous sensing modalities. To further improve long-term stability, MARVO applies a reinforcement-learning (RL) policy in order to refine the SE(2) pose graph; this is inspired by recent work in efficient pose-graph optimization~\cite{moreira2021fast}. Instead of solely relying on non-convex least-squares solvers which often get stuck in suboptimal minima, the learned policy proposes retraction actions to steer the trajectory towards globally consistent solutions. This learned refinement complements classical optimization and is particularly effective for turbid or visually sparse scenes.

MARVO integrates physics-informed perception, probabilistic fusion, and learning-based optimization into a reliable underwater localization system. Its contributions are:
\begin{itemize}
\item \textbf{Physics-aware front end:} a differentiable radiance adapter that compensates wavelength-dependent attenuation within the LoFTR transformer pipeline.
\item \textbf{Probabilistic multi-sensor fusion:} a visual–inertial–barometric estimator based on a GTSAM fixed-lag smoother with PARA-enhanced constraints.
\item \textbf{Learning-based global optimization:} an RL-driven pose-graph optimizer that refines SE(2) trajectories beyond the limits of standard least-squares solvers.
\end{itemize}

Together, these elements provide a single powerful underwater VO system that is able to estimate geometrically consistent trajectories where traditional pipelines fail.

\section{Related Works}\
\label{sec:related}
MARVO leverages three fundamental research directions: feature matching, factor-graph visual-inertial odometry, and reinforcement-learning-based pose-graph optimization.


\subsection{Feature Matching}\

\textbf{Detector-based methods.} Classical local features follow a paradigm of detect-describe-match. While handcrafted techniques such as SIFT~\cite{lowe2004sift}, SURF~\cite{bay2006surf}, and ORB~\cite{rublee2011orb} enjoy robustness to viewpoint and illumination changes, they fail in texturepoor or turbidity degraded underwater imagery. Similarly, learned techniques that improve repeatability via data-driven keypoint detection, such as LIFT~\cite{yi2016lift} and SuperPoint~\cite{detone2018superpoint}, are still limited by the fundamental sparsity of stable interest points in underwater scenes.

\textbf{Detector-free methods.} Dense matching methods avoid explicit keypoint detection and compute cost volumes or dense correlation fields directly. NCNet~\cite{rocco2018neighbourhood} and DRC-Net~\cite{li2020dual} impose neighborhood consensus on dense descriptors, although their convolutional receptive fields limit global reasoning. LoFTR~\cite{loftr} introduced an alternating mechanism of self- and cross-attention in establishing globally consistent semi-dense matches; since then, LoFTR has been widely adopted as a front-end for geometric vision tasks. However, descriptor quality degrades due to contrast loss, backscatter, and wavelength-dependent attenuation underwater, which motivated the physics-aware radiance adaptation in MARVO.

\textbf{Transformers in geometric vision.} Transformers have become foundational in correspondence and motion estimation due to their ability to model long-range relationships. Following ViT~\cite{dosovitskiy2020vit}, attention-based architectures have been deployed in optical flow~\cite{teed2020raft}, correspondence estimation~\cite{loftr}, and geometric reasoning. MARVO extends this paradigm with a Physics-Aware Radiance Adapter (PARA) that modulates LoFTR’s intermediate features according to learned attenuation cues, restoring discriminability in underwater conditions while retaining global attention benefits.


\subsection{Factor Graph-Based Visual Inertial Odometry}

Pose-graph and factor-graph estimation.
Robotic state estimation is often framed as a sparse nonlinear least-squares problem over poses and landmarks, modeled probabilistically with factor graphs~\cite{dellaert2017factorgraphs}. Solvers such as GTSAM~\cite{dellaert2017factorgraphs}, g2o~\cite{kummerle2011g}, and iSAM2~\cite{kaess2012isam2} leverage sparsity to enable real-time inference and form the basis of many state-of-the-art VO and SLAM methods. Visual–inertial odometry and preintegration. Combining cameras with IMUs leads to drift-resistant trajectories, assuming that the system properly models high-rate inertial signals. IMU preintegration~\cite{forster2017imu} allows for the efficient incorporation of continuous IMU measurements in fixed-lag smoothing frameworks by avoiding full re-integration during optimization. This has given rise to both highly accurate and computationally efficient VIO systems~\cite{leutenegger2015keyframe} suitable for real-time deployment. MARVO adheres to this paradigm, fusing PARA-enhanced visual constraints with IMU and barometric depth in a fixed-lag factor graph. In practical robotic systems, sensors run at different rates. Continuous-time fusion and multi-threaded architectures handle asynchronous data efficiently by interpolating trajectory states and marginalizing older poses~\cite{geneva2018asynchronous,lv2023continuous}. MARVO uses a similar strategy to enable real-time optimization with heterogeneous visual–inertial–pressure data. 

\subsection{Reinforcement-Learning-Based Pose-Graph Optimization} Classical PGO. Pose-graph optimization (PGO) solves for globally consistent trajectories by minimizing rotational and translational constraints over or, typically via Gauss-Newton or Levenberg-Marquardt with sparse factorizations. While effective, these solvers remain sensitive to poor initializations, unreliable loop closures, and strong noise-all common challenges in underwater environments. RL-based PGO. Recent work proposes reinforcement-learning-based optimizers that extend classical solvers by exploring the pose manifold beyond local gradients. RL-PGO~\cite{kourtzanidis2023rl} formulates planar PGO as a POMDP and learns retraction actions on \(\mathrm{SE}(2)\) that escape suboptimal minima. Distributed extensions further leverage graph neural networks for multi-robot settings~\cite{krishna2025policies}. MARVO follows a similar philosophy, using a learned policy to refine pose-graph estimates and enhance global consistency in the presence of underwater visual degradation.

\section{Dataset Augmentation}\
\label{sec:dataset}
To generate realistic underwater-appearance training data, we utilise the \textit{SyreaNet} synthesis module~\cite{syreanet}, which applies a physics-based underwater image formation model to in-air RGB-D datasets. We process RGB–depth pairs from ScanNet~\cite{dai2017scannet}, TartanAir~\cite{wang2020tartanair}, and Hypersim~\cite{roberts2021hypersim}, converting each into a synthetic underwater counterpart that simulates wavelength-dependent attenuation and depth-dependent scattering.

\begin{figure}[h]
\centering
\includegraphics[width=0.9\columnwidth]{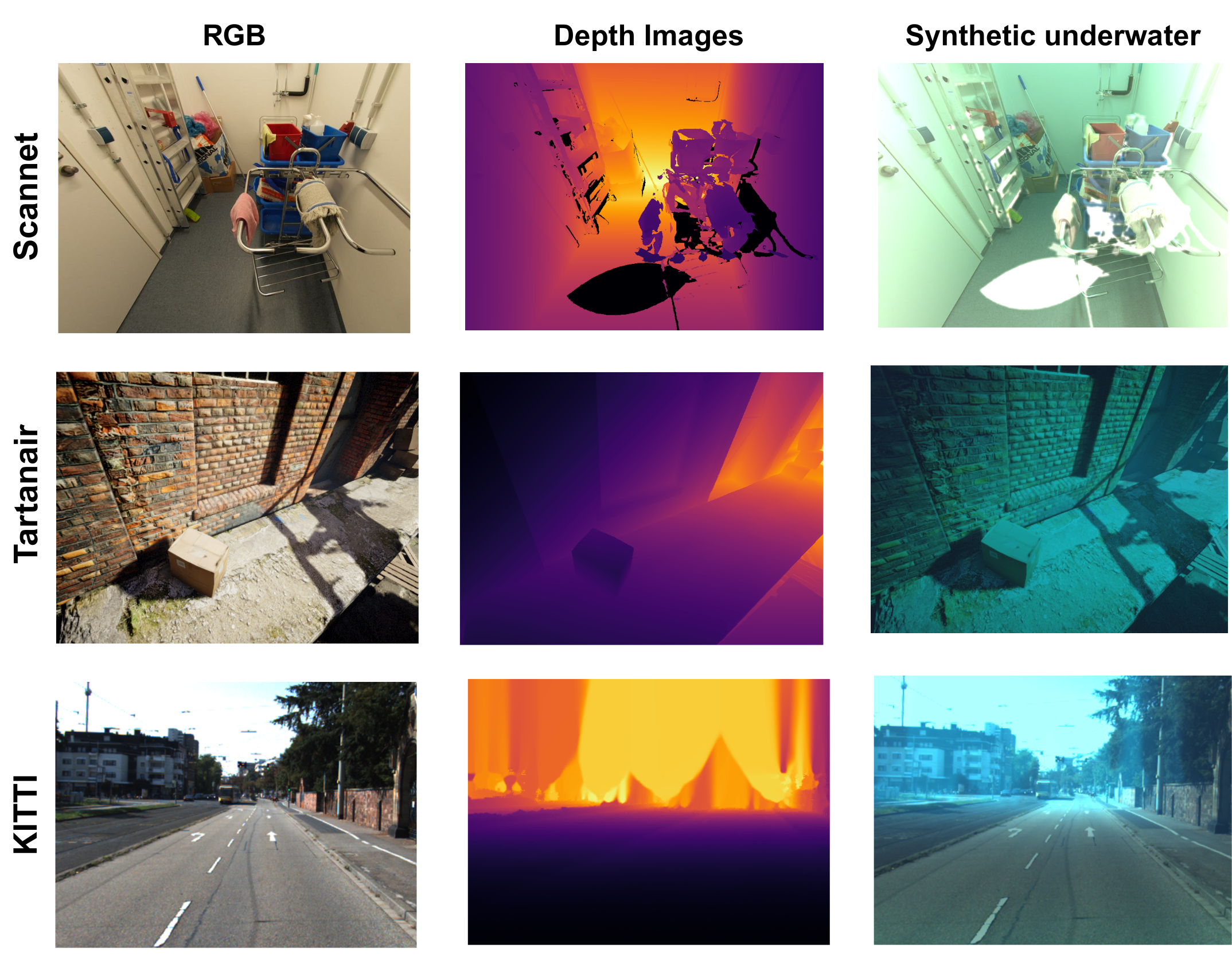}
\caption{Examples of synthetic data used in MARVO. Each row shows an example from a different dataset (ScanNet, TartanAir, Hypersim): original RGB image, corresponding depth map, and the generated synthetic underwater image with simulated attenuation and scattering.}
\label{fig:data_augmentation}
\end{figure}

\subsection{Physical Model}\

Following the revised underwater image formation model~\cite{akkaynak2018revised}, the observed intensity for color channel $c\!\in\!\{R,G,B\}$ is
\begin{equation}
I_c(x) = J_s(x)\, W_c \, e^{-\beta_c^D z(x)} + B_c(x),
\label{eq:formation}
\end{equation}
where $J_s(x)$ is the in-air scene radiance, $W_c$ is the diffuse downwelling term, $\beta_c^D$ is the attenuation coefficient, $B_c(x)$ is the backscatter component, and $z(x)$ is the per-pixel range. This formulation decouples attenuation and backscatter, enabling physically faithful simulation of haze, color loss, and contrast degradation under varying water types.

\subsection{Synthetic Generation}\

For every RGB–depth pair, SyreaNet samples $\beta_c^D$ and $B_c(x)$ from empirical distributions calibrated on real underwater imagery, emulating different levels of turbidity and spectral absorption. Illumination variability is emulated by stochastically perturbing $W_c$. In total, we synthesize approximately 120k RGB-D pairs, around 40k per dataset. We resize all images to $640{\times}480$ and normalize them to zero-mean, unit variance before training. The resulting renderings preserve the geometric structure of the original datasets but embed realistic radiometric degradation.

\subsection{Evaluation and Fine-Tuning}\

We asses the generalization of the front-end on KITTI~\cite{geiger2013vision} in a two-stage protocol: first, perform large-scale pretraining on synthetic underwater data; second, fine-tune on a small subset (10\%) of real data. This helps in bridging the synthetic-to-real domain gap while preserving the physical diversity of the synthetic corpus.

We further benchmark the feature-matching pipeline on MegaDepth~\cite{Li_2018_CVPR} using the standard image-pair splits and evaluation protocol. MegaDepth provides diverse Internet photo collections with SfM/MVS-derived geometry, allowing us to quantify improvements in match recall and pose accuracy under wide-baseline viewpoint and illumination changes.

\section{Feature Matching}
\label{sec:feature_matching}

At the heart of visual odometry lies feature correspondence. MARVO extends the detector-free matcher LoFTR~\cite{loftr} by adding a physics-guided radiance module inspired from SyreaNet~\cite{syreanet} for capturing the complex effects of attenuation, scattering, and color imbalance in underwater media. Combining them ensures geometric and photometric consistency even under wavelength-dependent distortions.

\begin{figure}[t]
    \centering
    \includegraphics[width=\columnwidth]{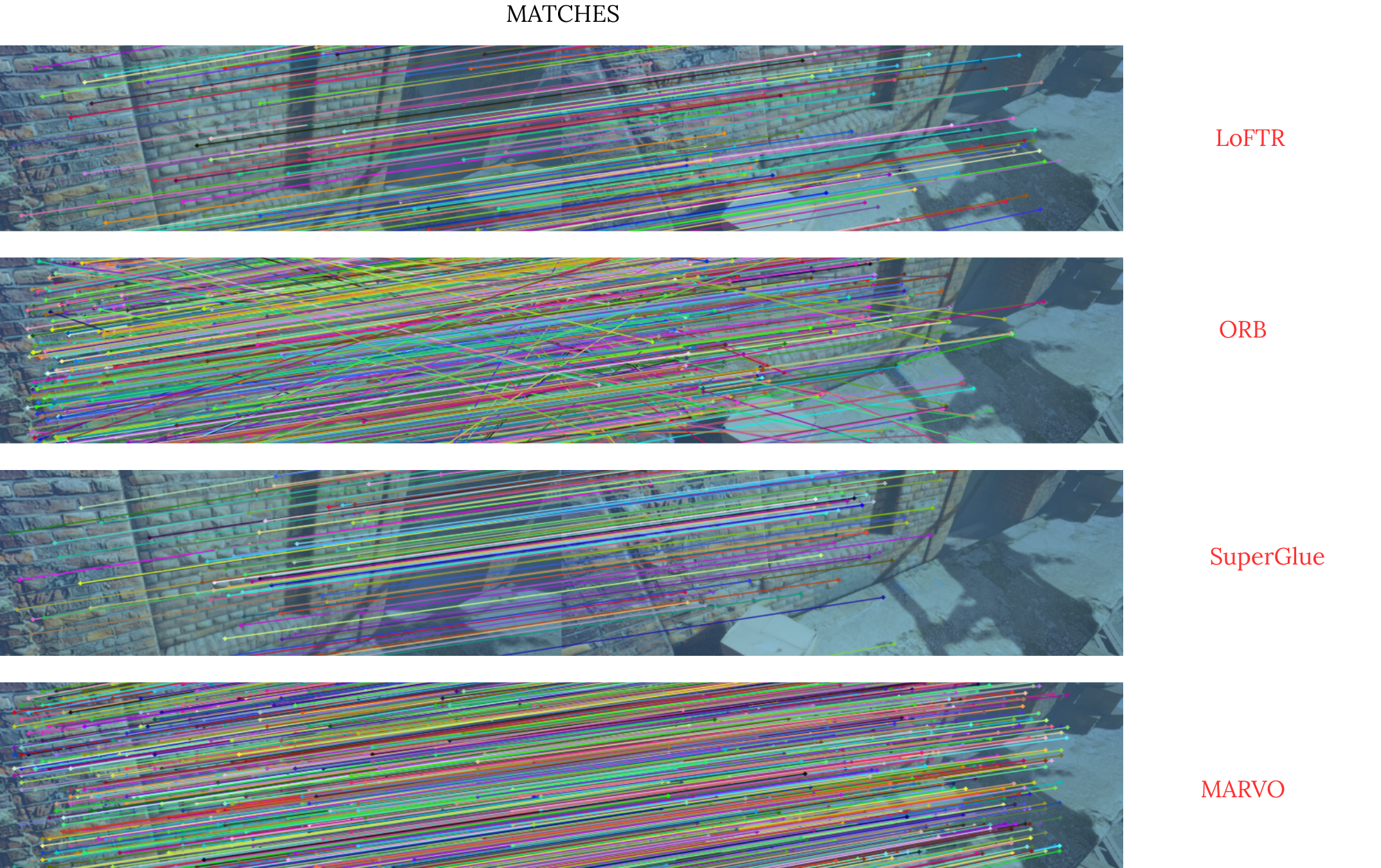}
    \caption{Qualitative feature matching comparison:
    MARVO produces denser and more geometrically stable correspondences than SuperGlue, ORB, and LoFTR under underwater conditions characterized by turbidity, color attenuation, and low texture. Conventional matchers degrade noticeably, while MARVO maintains semi-dense and spatially coherent matches through physics-aware radiance modulation.}
    \label{fig:feature_matching_comparison}
\end{figure}

\subsection{Detector-Free Transformer Backbone}

MARVO follows LoFTR's detector-free paradigm and directly predicts semi-dense correspondences from input image pairs with no explicit keypoint detection. Given rectified images $(I_A, I_B)$, convolutional encoders generate feature maps at $1/8$ resolution. The features are enriched with global context across both images by stacked self- and cross-attention layers. High-confidence correspondences are obtained from coarse correlation maps $\mathbf{C}_{\text{coarse}}$ using dual-softmax and mutual nearest-neighbor filtering and then refined to sub-pixel matches $\mathbf{C}_{\text{fine}}$ via differentiable correlation. This allows the detector-free architecture to perform robust matching in texture-poor underwater scenarios.

\subsection{Physics-Aware Radiance Adapter (PARA)}

To adapt transformer matching to underwater imaging, MARVO introduces a \textit{Physics-Aware Radiance Adapter} (PARA) between the CNN encoder and transformer layers.

\begin{figure}[t]
    \centering
    \includegraphics[width=\columnwidth]{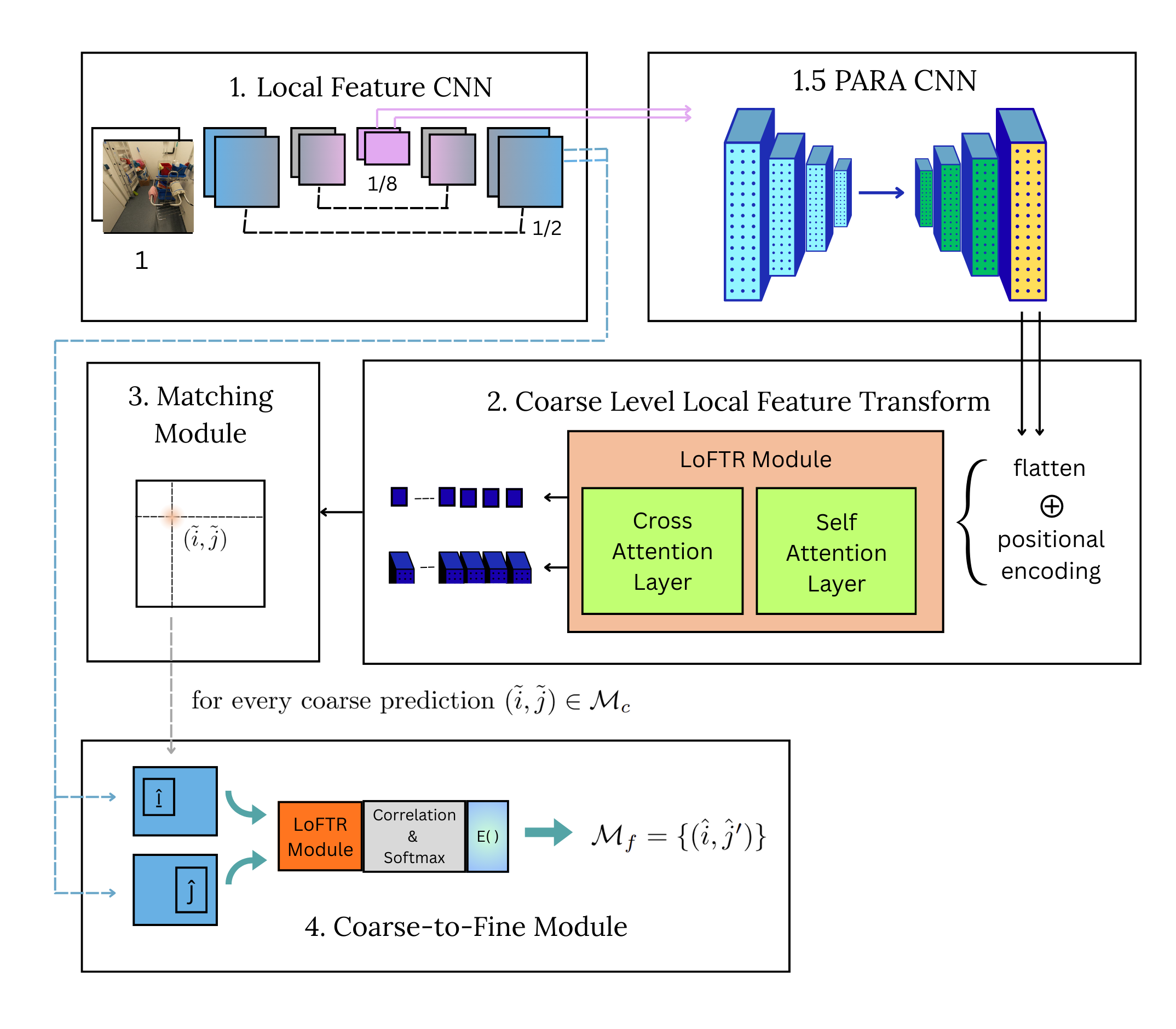}
    \caption{PARA architecture: The Physics-Aware Radiance Adapter takes coarse CNN features and predicts per-pixel attenuation and backscatter fields. These are used to generate a radiance correction mask $\Gamma(x)$, which normalizes intermediate descriptors before transformer matching. PARA compensates for wavelength-dependent attenuation, color imbalance, and contrast degradation common in underwater environments.}
    \label{fig:para_architecture}
\end{figure}

Let $\mathbf{F} \in \mathbb{R}^{H \times W \times C}$ denote the encoder feature map at $1/8$ resolution and let $\mathbf{I} \in \mathbb{R}^{H \times W \times 3}$ be the corresponding RGB image, both bilinearly downsampled to the same spatial size. PARA is implemented as a lightweight three-branch prediction head:
\begin{align}
\mathbf{F}_{\text{shared}} &= \phi_{\text{sh}}([\mathbf{F}, \text{BN}(\mathbf{I})]), \\
\hat{\boldsymbol{\beta}} &\in \mathbb{R}^{H \times W \times 3}
    = \phi_{\beta}(\mathbf{F}_{\text{shared}}), \\
\hat{\mathbf{B}}_{\infty} &\in \mathbb{R}^{H \times W \times 3}
    = \phi_{B}(\mathbf{F}_{\text{shared}}), \\
\hat{\mathbf{z}} &\in \mathbb{R}^{H \times W \times 1}
    = \phi_{z}(\mathbf{F}_{\text{shared}}),
\end{align}
where $\phi_{\text{sh}}$ denotes two $3\times3$ convolutional layers with ReLU and batch normalization, and $\phi_{\beta}, \phi_{B}, \phi_{z}$ are $1\times1$ convolutional heads that produce per-pixel attenuation, backscatter, and a depth proxy respectively. All predictions are made at the feature resolution, so no extra decoder is required.

We model the underwater image formation for each color channel $c \in \{R,G,B\}$ as
\begin{equation}
I_c(x) = J_c(x) \, e^{-\beta_c(x) z(x)}
        + B_\infty^c(x) \bigl(1 - e^{-\beta_c(x) z(x)}\bigr),
\label{eq:uw_formation}
\end{equation}
where $J_c(x)$ is the in-air radiance, $\beta_c(x)$ is the attenuation coefficient, $B_\infty^c(x)$ is the asymptotic backscatter, and $z(x)$ is the range. During training on synthetic RGB–depth data, we supervise $\hat{\boldsymbol{\beta}}$ and $\hat{\mathbf{B}}_{\infty}$ using the corresponding SyreaNet parameters and provide the ground-truth depth $z(x)$ to PARA. At test time, PARA relies only on the predicted proxy $\hat{\mathbf{z}}$.

To obtain a radiance-corrected estimate of $J_c(x)$ we invert Eq.~\eqref{eq:uw_formation} using the predicted fields:
\begin{equation}
\hat{J}_c(x) =
\bigl(I_c(x) - \hat{B}_\infty^c(x) (1 - e^{-\hat{\beta}_c(x) \hat{z}(x)})\bigr)
\cdot e^{\hat{\beta}_c(x) \hat{z}(x)}.
\label{eq:j_hat}
\end{equation}
From $\hat{J}_c(x)$ we derive a scalar, channel-aggregated correction mask
\begin{equation}
\Gamma(x) =
\frac{1}{3} \sum_{c \in \{R,G,B\}}
\frac{\hat{J}_c(x)}{I_c(x) + \epsilon},
\label{eq:gamma_mask}
\end{equation}
with a small $\epsilon$ to avoid division by zero. The mask $\Gamma(x) \in \mathbb{R}^{H \times W \times 1}$ is then broadcast across channels and applied to the encoder features as
\begin{equation}
\tilde{\mathbf{F}}(x) =
\text{LN}\bigl(\Gamma(x) \odot \mathbf{F}(x)\bigr),
\label{eq:feature_norm}
\end{equation}
where $\odot$ denotes element-wise multiplication and $\text{LN}(\cdot)$ is layer normalization. The transformer matcher in MARVO operates on $\tilde{\mathbf{F}}$ instead of $\mathbf{F}$, so all subsequent attention layers see features that have been explicitly corrected for wavelength-dependent attenuation and backscatter.

In practice, PARA adds fewer than $5\%$ additional parameters relative to the LoFTR backbone and keeps the feature resolution unchanged, which preserves LoFTR's computational profile while significantly improving descriptor consistency in spectrally distorted regions.

\subsection{Training Objectives}

The front-end is trained using a combined geometric, photometric, and physics-based loss:
\begin{align}
\mathcal{L} &=
\lambda_{\text{match}}\mathcal{L}_{\text{match}}
+ \lambda_{\text{photo}}\mathcal{L}_{\text{photo}}
+ \lambda_{\text{phys}}\mathcal{L}_{\text{phys}}, \\
\mathcal{L}_{\text{match}} &= \|\hat{\mathbf{P}} - \mathbf{P}^*\|_1, \\
\mathcal{L}_{\text{photo}} &= 1 - \text{SSIM}(I'_A, I'_B), \\
\mathcal{L}_{\text{phys}} &=
\|\hat{\boldsymbol{\beta}} - \boldsymbol{\beta}_{\text{gt}}\|_1 +
\|\hat{\mathbf{B}}_\infty - \mathbf{B}_{\infty,\text{gt}}\|_1.
\end{align}
Here $\boldsymbol{\beta}_{\text{gt}}$ and $\mathbf{B}_{\infty,\text{gt}}$ are the SyreaNet-derived supervision fields used in Eq.~\eqref{eq:uw_formation}. $\mathcal{L}_{\text{match}}$ enforces geometric consistency of correspondences, $\mathcal{L}_{\text{photo}}$ encourages radiance-corrected view agreement through the PARA-normalized images $I'_A, I'_B$, and $\mathcal{L}_{\text{phys}}$ explicitly ties the predicted physical fields to the underlying underwater image formation model.

\subsection{Synthetic-to-Real Adaptation}

MARVO is trained on a combination of synthetic underwater data and real underwater images. A physics-based synthesis pipeline generates pairs $(I_{\text{air}}, I_{\text{uw}})$ across multiple turbidity and illumination conditions. Domain adaptation and consistency regularization encourage feature invariance across synthetic-real shifts, thus allowing for robust matching without environment-specific calibration.

\subsection{Training Procedure}

Our model is trained in two stages: (1) pre-training with $1.2\times10^5$ synthetic underwater image pairs from ScanNet~\cite{dai2017scannet}, TartanAir~\cite{wang2020tartanair} and Hypersim~\cite{roberts2021hypersim}; and (2) fine-tuning with $\sim$12k real underwater frames including 10\% KITTI~\cite{geiger2013vision} and internal field data. The convolutional layers are partially frozen while fine-tuning transformer and PARA modules using real-world attenuation statistics. Training utilizes mixed-precision and multi-GPU parallelization on 4$\times$ NVIDIA A100 GPUs.

The developed physics-aware transformer recovers stable semi-dense matches even in spectrally distorted and feature-poor underwater regions, forming the primary constraints for downstream GTSAM-based state estimation and RL-PGO global optimization.

\section{Pose Graph Localization}
\label{sec:localization}

State estimation in MARVO is done by a lightweight factor graph that fuses: i) physics-aware visual constraints from PARA–LoFTR, ii) IMU preintegration, and iii) a barometric depth prior. A fixed-lag smoother is implemented for the back-end in GTSAM. Unlike typical VIO pipelines, MARVO involves two new components specific to underwater applications-a semi-dense visual factor derived from physics-corrected matches, and a unary depth factor that suppresses vertical drift.

\subsection{State Representation}

Every keyframe state $\mathbf{x}_i$ contains orientation, position, velocity, and IMU biases. Only keyframes remain in the optimization window for real-time operation.

\begin{figure*}[t]
    \centering
    \includegraphics[width=\textwidth]{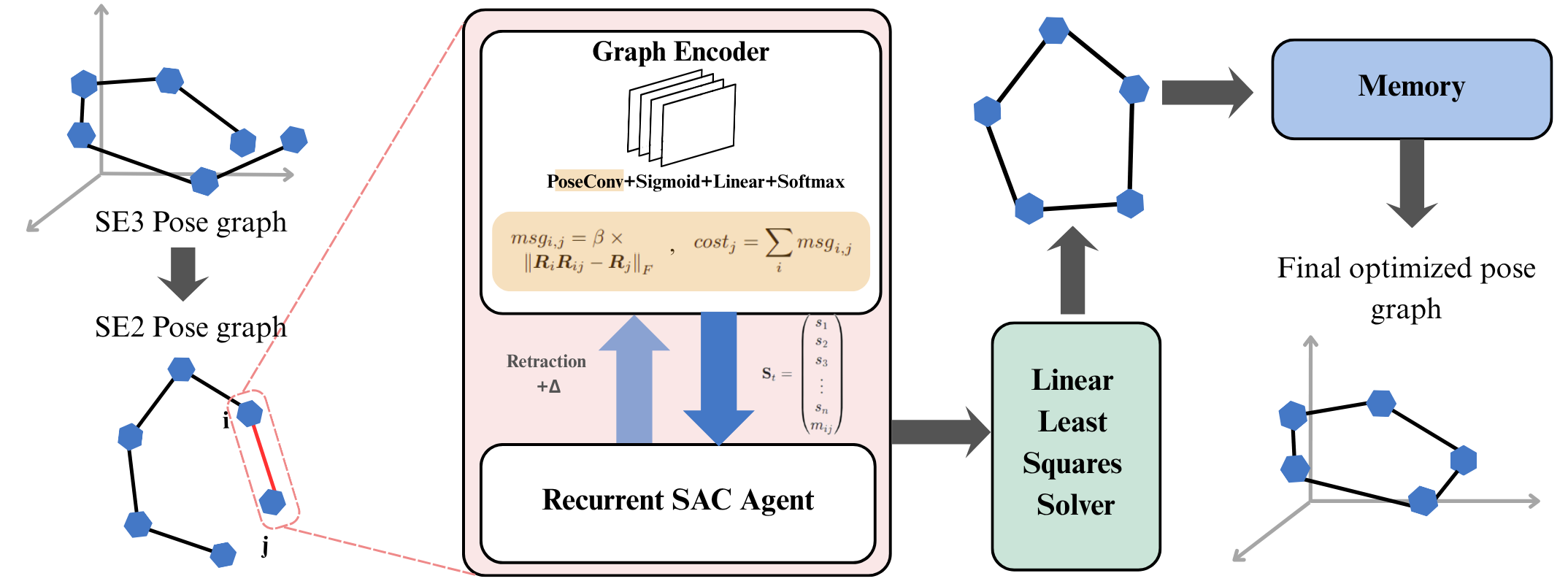}
    \caption{Reinforcement learning-based pose-graph optimization. A GNN encoder maps an initial pose graph to latent edge features, which condition a recurrent SAC agent. The agent iteratively applies retraction actions; a final linear least-squares step produces the optimized pose graph.}
    \label{fig:architecture}
\end{figure*}

\subsection{Graph Factors}

\textbf{IMU Preintegration.}
We apply standard GTSAM preintegration to provide metric scale and short-term motion constraints.

\textbf{Depth Prior (Underwater).}
A barometric pressure reading provides a explicit estimate of depth. We add a simple unary factor on $\mathbf{p}_i$ an inexpensive but highly effective constraint that eliminates the vertical drift common in monocular and low-visibility underwater VO.

\textbf{MARVO Visual Factor.}
For every keyframe pair $(i,j)$, PARA–LoFTR provides semi-dense matches
\[
\mathcal{M}_{ij} = \{(\mathbf{u}^i_k,\,\mathbf{u}^j_k)\}_{k=1}^{N},
\]
from which we estimate a relative pose via an essential matrix. A scale variable $s_{ij}$ is co-optimized with the poses, improving robustness under partial stereo loss or poor geometric parallax.

\textbf{Adaptive Covariance.}
MARVO weights each visual factor using physics-aware match confidences. The covariance scales inversely with inlier count and spatial coverage, allowing high-visibility frames to dominate while  downweighting degraded or spectrally imbalanced images automatically .

\smallskip
MARVO’s localization back-end differs from conventional VIO in two aspects: (1) PARA–LoFTR supplies stable, semi-dense constraints even under turbidity and color attenuation, and (2) an underwater-specific depth factor provides drift-free vertical motion. These components produce a reliable initialization for the RL-based global refinement stage that follows.

\section{Offline Agent-Initiated Pose Graph Optimisation}

\subsection{Foundations}
Classical iterative solvers for pose-graph optimization, such as Gauss--Newton and Levenberg--Marquardt \cite{kummerle2011g, dellaert2017factorgraphs}, are efficient but prone to local minima under poor initialization or heavy noise. RL-PGO \cite{kourtzanidis2023rl} addresses this by casting pose-graph optimization as a partially observable Markov decision process on $\mathrm{SE}(2)$ and learning a policy that applies continuous retraction actions to pose pairs. A message-passing GNN encodes edge residuals \cite{Azzam2021PoseGraph}, enabling the agent to explore beyond local gradients and systematically bootstrap classical solvers on graphs of varying size.

\subsection{Domain Adaptation for Autonomous Underwater Craft}
AUVs and ROVs are often roll/pitch stabilized via ballast and thrusters, leaving yaw as the main rotational degree of freedom, while depth is accurately measured by a pressure sensor \cite{ forster2017imu}. This motivates a dimensionality-reduced formulation: from the GTSAM-based visual–inertial–barometric frontend \cite{dellaert2012factor, kaess2012isam2} we extract the planar $\mathrm{SE}(2)$ states $(x,y,\theta)$, build a 2D pose graph on the horizontal plane, and keep roll, pitch, and depth fixed via barometric priors.

\subsection{Architecture and Integration}
Our optimizer follows a two-stage pipeline. First, the full $\mathrm{SE}(3)$ factor graph is optimized with a standard solver (iSAM2 or Levenberg--Marquardt) \cite{kaess2012isam2, dellaert2017factorgraphs} to obtain a consistent initial trajectory. We then project this trajectory to $\mathrm{SE}(2)$ and feed the planar graph to the RL agent for refinement. After convergence, the corrected $(x,y,\theta)$ poses are lifted back to $\mathrm{SE}(3)$ by reattaching the original roll, pitch, and depth estimates, yielding a globally consistent 3D trajectory.

\textbf{Graph Encoding and Message Passing:}
We use a GNN encoder that aggregates orientation and translation residuals on each edge $(i,j)\in E$ \cite{kourtzanidis2023rl, Azzam2021PoseGraph}. Message passing yields per-edge embeddings capturing patterns such as odometric chains and loop closures; these embeddings are pooled into the state fed to the policy.

\textbf{Policy and Sequential Refinement:}
A recurrent Soft Actor–Critic (SAC) agent with LSTM history \cite{haarnoja2018sac} selects an edge and outputs a retraction action in $\mathrm{SE}(2)$ at each step. Retractions are applied via the exponential map \cite{blanco2021tutorial}, enabling smooth, continuous pose updates while preserving manifold structure.

\subsection{Key Innovation: Log-Weighted Orientation Cost}
RL-PGO averages orientation errors uniformly over all edges \cite{kourtzanidis2023rl}, implicitly assuming equal importance. We instead weight each edge’s rotational error by the magnitude of its associated translation, yielding the log-weighted orientation cost
\begin{equation}
    OC_{\text{log}} = \sqrt{\sum_{(i,j) \in E} w_{ij} \left\| R_i R_{ij} - R_j \right\|_F^2},
    \label{eq:oc_log}
\end{equation}
with
\begin{equation}
    w_{ij} = 1 + \beta \log\!\left( \frac{\| \mathbf{t}_{ij} \|}{\bar{t}} + \epsilon \right),
    \label{eq:log_weight}
\end{equation}
where $R_i,R_j,R_{ij}\in\mathrm{SO}(2)$ are absolute and relative rotations, $\mathbf{t}_{ij}\in\mathbb{R}^2$ is the measured translation, $\bar{t}$ is the mean translation magnitude over all edges, $\beta$ controls the strength of the weighting, and $\epsilon$ is a small constant for numerical stability.

The logarithmic form remains monotonically increasing but sublinear in $\|\mathbf{t}_{ij}\|$: long-range constraints are emphasized without allowing a few very long, noisy edges to dominate the reward. Setting $\beta=0$ recovers the uniform weighting of \cite{kourtzanidis2023rl}, while moderate $\beta$ values better reflect the heterogeneous uncertainty patterns typical in underwater SLAM.

\subsection{Post-Optimization Expansion and Final Refinement}
After RL-based $\mathrm{SE}(2)$ refinement, we reattach roll, pitch, and depth from the initial GTSAM estimate (or barometric priors) to obtain a full $\mathrm{SE}(3)$ trajectory. A final short Levenberg--Marquardt refinement \cite{kummerle2011g, moreira2021fast} on the reconstructed 3D pose graph exploits the improved initialization to quickly converge to a high-quality local minimum, even in regimes where purely classical methods tend to stall \cite{carlone2015pgo}.

\section{Experiments}
\label{sec:experiments}

We evaluate MARVO across synthetic underwater benchmarks and real coastal field deployments, comparing against both classical and learning-based VO systems. Our experiments measure (1) correspondence quality under wavelength-dependent attenuation and (2) end-to-end odometry accuracy. Due to the absence of standardized underwater VO datasets with ground-truth poses and depth maps, our evaluation incorporates both physically rendered datasets and real sequences aligned to SfM-based ground truth. This limitation is inherent to underwater benchmarking: accurate supervision requires synchronized RGB, depth, and high-quality pose annotations, which are rarely available together in public datasets.

\subsection{Evaluations}

\textbf{Feature Matching Accuracy.}
We first benchmark MARVO's radiance-aware correspondence module against SuperGlue and LoFTR~\cite{loftr}. Using RGB-D datasets rendered through a physically based underwater model (per-channel attenuation and scattering), we compute the pose estimation AUC at $5^\circ$, $10^\circ$, and $20^\circ$. MARVO achieves the highest accuracy across all thresholds, showing improved match stability under spectral degradation.

\begin{table}[h]
\centering
\caption{Pose estimation AUC on synthetic underwater sequences.}
\label{tab:pose_estimation}
\begin{tabular}{lccc}
\toprule
Method & 5° & 10° & 20° \\
\midrule
SP + SuperGlue & 25.4 & 42.2 & 59.7 \\
LoFTR & 42.9 & 59.5 & 68.2 \\
\textbf{MARVO (Ours)} & \textbf{49.7} & \textbf{62.9} & \textbf{71.3} \\
\bottomrule
\end{tabular}
\end{table}

\textbf{Visual Odometry on Synthetic Underwater Dataset.}
We next evaluate end-to-end VO performance. All trajectories are scale-aligned using a similarity transform. MARVO substantially reduces ATE, angular drift, and relative pose error compared to classical baselines and modern VO pipelines.

\begin{table}[h]
\centering
\small
\caption{Synthetic underwater VO performance}
\label{tab:our_data_vo}
\begin{tabularx}{\columnwidth}{lXXX}
\toprule
Method & ATE (m) $\downarrow$ & RPE (deg/m) $\downarrow$ & Drift (\%) $\downarrow$ \\
\midrule
ORB-SLAM3 & 6.45 & 1.38 & 5.9 \\
LIBVISO2 & 5.12 & 1.14 & 5.1 \\
\textbf{MARVO (ours)} & \textbf{2.47} & \textbf{0.61} & \textbf{2.2} \\
\bottomrule
\end{tabularx}
\end{table}

\textit{Note:} Larger-scale multi-sequence evaluations are limited by the scarcity of datasets that provide jointly calibrated ground truth depth, images, and poses, all of which are required to generate physically realistic underwater renderings.

\subsection{Real-World Field Data}
We evaluate MARVO on real underwater imagery collected using a monocular camera, IMU, and depth-pressure sensors. Ground truth camera poses are obtained via COLMAP SfM and aligned using a 7-DoF similarity transform. These sequences contain severe turbidity, forward-scattering, and intermittent visibility. MARVO maintains significantly lower drift and angular error compared to classical VO methods, validating its robustness in challenging real-world conditions.

\begin{table}[h]
\centering
\small
\caption{VO performance on real underwater field deployments (Scale Alligned)}
\label{tab:field_data_vo}
\begin{tabularx}{\columnwidth}{lXXX}
\toprule
Method & ATE (m) $\downarrow$ & RPE (deg/m) $\downarrow$ & Drift (\%) $\downarrow$ \\
\midrule
ORB-SLAM3 & 4.12 & 0.92 & 3.8 \\
LIBVISO2 & 3.47 & 0.85 & 3.1 \\
MAST3R-SLAM \cite{murai2024_mast3rslam} & 2.52 & 0.58 & 2.2 \\
VGGT-SLAM \cite{maggio2025vggtslamdensergbslam} & 2.41 & 0.56 & 2.1 \\
\textbf{MARVO (ours)} & \textbf{1.73} & \textbf{0.34} & \textbf{1.2} \\
\bottomrule
\end{tabularx}
\end{table}

\subsection{Ablation Studies}
\label{subsec:ablations}

We ablate MARVO to quantify the contribution of each module:

\begin{itemize}
    \item \textbf{No PARA Module:} Removing physics-guided radiance correction degrades matchability and increases drift, confirming its necessity under spectral attenuation.
    \item \textbf{Replace Matcher with LoFTR:} Using vanilla LoFTR instead of PARA-enhanced descriptors produces significantly worse correspondence reliability.
    \item \textbf{Replace RL-PGO with Classical PGO:} Gauss–Newton optimization leads to higher drift, especially when loop closures are sparse.
    \item \textbf{No Physics-Based Radiance Norm:} Using PARA without physics-based normalization yields the largest AUC drop, indicating that physics supervision—not merely CNN modulation—is responsible for robustness.
\end{itemize}

\begin{table}[h]
\centering
\small
\caption{Ablation analysis of MARVO components.}
\label{tab:feature_matching}
\begin{tabularx}{\columnwidth}{lXXX}
\toprule
Configuration & AUC @10° $\uparrow$ & ATE (m) $\downarrow$ & Drift (\%) $\downarrow$ \\
\midrule
Full MARVO (ours) & \textbf{0.92} & \textbf{1.73} & \textbf{1.2} \\
No PARA Module & 0.81 & 2.24 & 1.9 \\
Replace Feature Matcher (LoFTR) & 0.76 & 2.47 & 2.3 \\
Replace RL-PGO w/ Classical PGO & 0.84 & 2.05 & 1.7 \\
No Physics-Based Radiance Norm & 0.73 & 2.68 & 2.6 \\
\bottomrule
\end{tabularx}
\end{table}

\section{Conclusion}

We introduced MARVO, a marine-adaptive visual odometry framework that integrates physics-guided feature correction, multi-sensor factor-graph estimation, and reinforcement-learned pose-graph optimization. By embedding a wavelength-dependent attenuation and backscatter model directly into the transformer correspondence pipeline, MARVO restores descriptor discriminability in visually degraded underwater scenes. Combined with an RL-enhanced global optimizer and visual–inertial–barometric frontend, MARVO produces stable, geometrically consistent trajectories across a wide spectrum of conditions.

Our evaluations demonstrate consistent improvements over SuperGlue, LoFTR~\cite{loftr}, ORB-SLAM3, and LIBVISO2 across AUC, ATE, RPE, and drift metrics, both on synthetic underwater renderings and real-world field deployments. MARVO maintains tracking where classical detectors fail and significantly reduces global trajectory drift through RL-based refinement.

\textbf{Limitations and Future Work.}
A major limitation is the lack of large-scale underwater VO datasets with ground-truth trajectories and depth maps. Physics-based rendering requires metric depth and camera calibration, and accurate odometry evaluation requires high-quality ground truth—two properties rarely available simultaneously. This limits the scope of quantitative benchmarks and necessitates hybrid evaluation using synthetic renderings and COLMAP-aligned real data.

Future extensions include joint 3D mapping (e.g., TSDF fusion or underwater-adapted MVS), learning full $\mathrm{SE}(3)$ global optimization with roll/pitch coupling, and integrating acoustic depth priors to handle extreme turbidity or complete visual dropout.

{
    \small
    \bibliographystyle{ieeenat_fullname}
    \bibliography{references}

@inproceedings{loftr,
  title={LoFTR: Detector-free local feature matching with transformers},
  author={Sun, Jiaming and Shen, Zehong and Wang, Yuang and Bao, Hujun and Zhou, Xiaowei},
  booktitle={Proceedings of the IEEE/CVF conference on computer vision and pattern recognition},
  pages={8922--8931},
  year={2021}
}

@inproceedings{moreira2021fast,
  title={Fast pose graph optimization via Krylov-Schur and Cholesky factorization},
  author={Moreira, Gabriel and Marques, Manuel and Costeira, Joao Paulo},
  booktitle={Proceedings of the IEEE/CVF Winter Conference on Applications of Computer Vision},
  pages={1898--1906},
  year={2021}
}

@article{forster2017imu,
  title={On-manifold preintegration for real-time visual--inertial odometry},
  author={Forster, Christian and Carlone, Luca and Dellaert, Frank and Scaramuzza, Davide},
  journal={IEEE Transactions on Robotics},
  volume={33},
  number={1},
  pages={1--21},
  year={2016},
  publisher={IEEE}
}

@article{dellaert2017factorgraphs,
  title={Factor graphs for robot perception},
  author={Dellaert, Frank and Kaess, Michael and others},
  journal={Foundations and Trends{\textregistered} in Robotics},
  volume={6},
  number={1-2},
  pages={1--139},
  year={2017},
  publisher={Now Publishers, Inc.}
}

@article{lowe2004sift,
  title={Distinctive image features from scale-invariant keypoints},
  author={Lowe, David G},
  journal={International journal of computer vision},
  volume={60},
  number={2},
  pages={91--110},
  year={2004},
  publisher={Springer}
}

@inproceedings{bay2006surf,
  title={Surf: Speeded up robust features},
  author={Bay, Herbert and Tuytelaars, Tinne and Van Gool, Luc},
  booktitle={European conference on computer vision},
  pages={404--417},
  year={2006},
  organization={Springer}
}

@inproceedings{rublee2011orb,
  title={ORB: An efficient alternative to SIFT or SURF},
  author={Rublee, Ethan and Rabaud, Vincent and Konolige, Kurt and Bradski, Gary},
  booktitle={2011 International conference on computer vision},
  pages={2564--2571},
  year={2011},
  organization={Ieee}
}

@inproceedings{yi2016lift,
  title={Lift: Learned invariant feature transform},
  author={Yi, Kwang Moo and Trulls, Eduard and Lepetit, Vincent and Fua, Pascal},
  booktitle={European conference on computer vision},
  pages={467--483},
  year={2016},
  organization={Springer}
}

@inproceedings{detone2018superpoint,
  title={Superpoint: Self-supervised interest point detection and description},
  author={DeTone, Daniel and Malisiewicz, Tomasz and Rabinovich, Andrew},
  booktitle={Proceedings of the IEEE conference on computer vision and pattern recognition workshops},
  pages={224--236},
  year={2018}
}

@inproceedings{kummerle2011g,
  title={g 2 o: A general framework for graph optimization},
  author={K{\"u}mmerle, Rainer and Grisetti, Giorgio and Strasdat, Hauke and Konolige, Kurt and Burgard, Wolfram},
  booktitle={2011 IEEE international conference on robotics and automation},
  pages={3607--3613},
  year={2011},
  organization={IEEE}
}

@article{rocco2018neighbourhood,
  title={Neighbourhood consensus networks},
  author={Rocco, Ignacio and Cimpoi, Mircea and Arandjelovi{\'c}, Relja and Torii, Akihiko and Pajdla, Tomas and Sivic, Josef},
  journal={Advances in neural information processing systems},
  volume={31},
  year={2018}
}

@article{li2020dual,
  title={Dual-resolution correspondence networks},
  author={Li, Xinghui and Han, Kai and Li, Shuda and Prisacariu, Victor},
  journal={Advances in Neural Information Processing Systems},
  volume={33},
  pages={17346--17357},
  year={2020}
}

@article{dosovitskiy2020vit,
  title={An image is worth 16x16 words: Transformers for image recognition at scale},
  author={Dosovitskiy, Alexey},
  journal={arXiv preprint arXiv:2010.11929},
  year={2020}
}

@inproceedings{teed2020raft,
  title={Raft: Recurrent all-pairs field transforms for optical flow},
  author={Teed, Zachary and Deng, Jia},
  booktitle={European conference on computer vision},
  pages={402--419},
  year={2020},
  organization={Springer}
}

@article{kourtzanidis2023rl,
  title={Rl-pgo: Reinforcement learning-based planar pose-graph optimization},
  author={Kourtzanidis, Nikolaos and Saeedi, Sajad},
  journal={IEEE Control Systems Letters},
  volume={7},
  pages={3777--3782},
  year={2023},
  publisher={IEEE}
}

@article{krishna2025policies,
  title={Policies over Poses: Reinforcement Learning based Distributed Pose-Graph Optimization for Multi-Robot SLAM},
  author={Krishna Ghanta, Sai and Parasuraman, Ramviyas},
  journal={arXiv e-prints},
  pages={arXiv--2510},
  year={2025}
}

@article{leutenegger2015keyframe,
  title={Keyframe-based visual--inertial odometry using nonlinear optimization},
  author={Leutenegger, Stefan and Lynen, Simon and Bosse, Michael and Siegwart, Roland and Furgale, Paul},
  journal={The International Journal of Robotics Research},
  volume={34},
  number={3},
  pages={314--334},
  year={2015},
  publisher={SAGE Publications Sage UK: London, England}
}

@inproceedings{geneva2018asynchronous,
  title={Asynchronous multi-sensor fusion for 3D mapping and localization},
  author={Geneva, Patrick and Eckenhoff, Kevin and Huang, Guoquan},
  booktitle={2018 IEEE international conference on robotics and automation (ICRA)},
  pages={5994--5999},
  year={2018},
  organization={IEEE}
}

@article{lv2023continuous,
  title={Continuous-time fixed-lag smoothing for LiDAR-inertial-camera SLAM},
  author={Lv, Jiajun and Lang, Xiaolei and Xu, Jinhong and Wang, Mengmeng and Liu, Yong and Zuo, Xingxing},
  journal={IEEE/ASME Transactions on Mechatronics},
  volume={28},
  number={4},
  pages={2259--2270},
  year={2023},
  publisher={IEEE}
}

@article{kaess2012isam2,
  title={iSAM2: Incremental smoothing and mapping using the Bayes tree},
  author={Kaess, Michael and Johannsson, Hordur and Roberts, Richard and Ila, Viorela and Leonard, John J and Dellaert, Frank},
  journal={The International Journal of Robotics Research},
  volume={31},
  number={2},
  pages={216--235},
  year={2012},
  publisher={Sage Publications Sage UK: London, England}
}

@article{carlone2015pgo,
  title={Pose graph optimization in the complex domain: Lagrangian duality, conditions for zero duality gap, and optimal solutions},
  author={Calafiore, Giuseppe and Carlone, Luca and Dellaert, Frank},
  journal={arXiv preprint arXiv:1505.03437},
  year={2015}
}

@article{dellaert2012factor,
  title={Factor graphs and GTSAM: A hands-on introduction},
  author={Dellaert, Frank},
  journal={Georgia Institute of Technology, Tech. Rep},
  volume={2},
  number={4},
  year={2012}
}

@article{syreanet,
  title={Syreanet: A physically guided underwater image enhancement framework integrating synthetic and real images},
  author={Wen, Junjie and Cui, Jinqiang and Zhao, Zhenjun and Yan, Ruixin and Gao, Zhi and Dou, Lihua and Chen, Ben M},
  journal={arXiv preprint arXiv:2302.08269},
  year={2023}
}

@inproceedings{dai2017scannet,
  title={Scannet: Richly-annotated 3d reconstructions of indoor scenes},
  author={Dai, Angela and Chang, Angel X and Savva, Manolis and Halber, Maciej and Funkhouser, Thomas and Nie{\ss}ner, Matthias},
  booktitle={Proceedings of the IEEE conference on computer vision and pattern recognition},
  pages={5828--5839},
  year={2017}
}

@inproceedings{wang2020tartanair,
  title={Tartanair: A dataset to push the limits of visual slam},
  author={Wang, Wenshan and Zhu, Delong and Wang, Xiangwei and Hu, Yaoyu and Qiu, Yuheng and Wang, Chen and Hu, Yafei and Kapoor, Ashish and Scherer, Sebastian},
  booktitle={2020 IEEE/RSJ International Conference on Intelligent Robots and Systems (IROS)},
  pages={4909--4916},
  year={2020},
  organization={IEEE}
}

@inproceedings{roberts2021hypersim,
  title={Hypersim: A photorealistic synthetic dataset for holistic indoor scene understanding},
  author={Roberts, Mike and Ramapuram, Jason and Ranjan, Anurag and Kumar, Atulit and Bautista, Miguel Angel and Paczan, Nathan and Webb, Russ and Susskind, Joshua M},
  booktitle={Proceedings of the IEEE/CVF international conference on computer vision},
  pages={10912--10922},
  year={2021}
}

@article{geiger2013vision,
  title={Vision meets robotics: The kitti dataset},
  author={Geiger, Andreas and Lenz, Philip and Stiller, Christoph and Urtasun, Raquel},
  journal={The international journal of robotics research},
  volume={32},
  number={11},
  pages={1231--1237},
  year={2013},
  publisher={Sage Publications Sage UK: London, England}
}

@inproceedings{akkaynak2018revised,
  title={A revised underwater image formation model},
  author={Akkaynak, Derya and Treibitz, Tali},
  booktitle={Proceedings of the IEEE conference on computer vision and pattern recognition},
  pages={6723--6732},
  year={2018}
}

@article{Azzam2021PoseGraph,
  title={Pose-graph neural network classifier for global optimality prediction in 2D SLAM},
  author={Azzam, Rana and Kong, Felix H and Taha, Tarek and Zweiri, Yahya},
  journal={IEEE Access},
  volume={9},
  pages={80466--80477},
  year={2021},
  publisher={IEEE}
}

@inproceedings{Li_2018_CVPR,
  title={Megadepth: Learning single-view depth prediction from internet photos},
  author={Li, Zhengqi and Snavely, Noah},
  booktitle={Proceedings of the IEEE conference on computer vision and pattern recognition},
  pages={2041--2050},
  year={2018}
}

@inproceedings{murai2024_mast3rslam,
  title={MASt3R-SLAM: Real-time dense SLAM with 3D reconstruction priors},
  author={Murai, Riku and Dexheimer, Eric and Davison, Andrew J},
  booktitle={Proceedings of the Computer Vision and Pattern Recognition Conference},
  pages={16695--16705},
  year={2025}
}

@article{maggio2025vggtslamdensergbslam,
  title={Vggt-slam: Dense rgb slam optimized on the sl (4) manifold},
  author={Maggio, Dominic and Lim, Hyungtae and Carlone, Luca},
  journal={arXiv preprint arXiv:2505.12549},
  year={2025}
}

@inproceedings{haarnoja2018sac,
  title={Soft actor-critic: Off-policy maximum entropy deep reinforcement learning with a stochastic actor},
  author={Haarnoja, Tuomas and Zhou, Aurick and Abbeel, Pieter and Levine, Sergey},
  booktitle={International conference on machine learning},
  pages={1861--1870},
  year={2018},
  organization={Pmlr}
}

@article{blanco2021tutorial,
  title={A tutorial on $\mathbf{SE}(3)$ transformation parameterizations and on-manifold optimization},
  author={Blanco-Claraco, José Luis},
  journal={arXiv preprint arXiv:2103.15980},
  year={2021}
}
}


\end{document}